\documentclass[runningheads]{llncs}

\usepackage[final,year=2026,ID=11587]{eccv}

\usepackage{eccvabbrv}

\usepackage{graphicx}
\usepackage{booktabs}
\usepackage{bm}

\usepackage[accsupp]{axessibility}  

\usepackage[hidelinks]{hyperref}

\usepackage{orcidlink}
\usepackage[most]{tcolorbox}
\usepackage{pifont}
\definecolor{percolor}{RGB}{40,90,180}
\definecolor{reacolor}{RGB}{180,60,40}
\usepackage{siunitx}
\definecolor{myorange}{RGB}{80, 101, 142}
\definecolor{lightblue}{RGB}{91, 182, 220}
\hypersetup{
    colorlinks=true,
    linkcolor=red,
    filecolor=myorange,
    urlcolor=lightblue,
    citecolor=lightblue,
}
\usepackage[table]{xcolor}
\definecolor{red}{RGB}{192,0,0}
\definecolor{blue}{RGB}{48, 85, 151}

\begin{document}

\title{Beyond Where to Look: Trajectory-Guided Reinforcement Learning for Multimodal RLVR}

\titlerunning{Trajectory-Guided Reinforcement Learning for Multimodal RLVR}

\author{
Jinda Lu\inst{1} \and
Junkang Wu\inst{1} \and
Jinghan Li\inst{1} \and
Kexin Huang\inst{1} \and
Shuo Yang\inst{2} \and \\
Mingzhu Chen\inst{1} \and
Jiancan Wu\inst{1} \and
Kuien Liu\inst{3} \and
Xiang Wang\inst{1}
}


\authorrunning{J.~Lu et al.}

\institute{University of Science and Technology of China\and
Pecking University \and
 Institute of Software, Chinese Academy of Sciences}

\maketitle

\begin{abstract}
Recent advances in Reinforcement Learning with Verifiable Rewards (RLVR) for multimodal large language models (MLLMs) have mainly focused on improving final answer correctness and strengthening visual grounding. 
However, a critical bottleneck remains: although models can attend to relevant visual regions, they often fail to effectively incorporate visual evidence into subsequent reasoning, leading to reasoning chains that are weakly grounded in visual facts. 
To address this issue, we propose \textbf{T}rajectory-\textbf{G}uided \textbf{R}einforcement \textbf{L}earning (TGRL), which guides the policy model to integrate visual evidence into fine-grained reasoning processes using expert reasoning trajectories from stronger models. 
We further introduce token-level reweighting and trajectory filtering to ensure stable and effective policy optimization. 
Extensive experiments on multiple multimodal reasoning benchmarks demonstrate that TGRL consistently improves reasoning performance and effectively bridges the gap between visual perception and logical reasoning.
\end{abstract}
\section{Introduction}
\label{sec:intro}
Reinforcement Learning with Verifiable Rewards (RLVR) has recently emerged as an effective paradigm for enhancing the reasoning capabilities of large language models (LLMs)~\cite{tulu3, qwen3, deepseekmath}. 
By leveraging automatically verifiable outcomes as reward signals, RLVR enables complex reasoning behaviors such as chain-of-thought (CoT) reasoning and self-reflection~\cite{gemini-2.5, openai-o1, deepseek_r1}. 
Building on these advances, recent studies extend RLVR to Multimodal Large Language Models (MLLMs), aiming to improve reasoning over visual inputs in multimodal tasks~\cite{perception_r1, vision_r1, visual_rft}.

\begin{figure}[t]
    \centering
    \includegraphics[width=\linewidth]{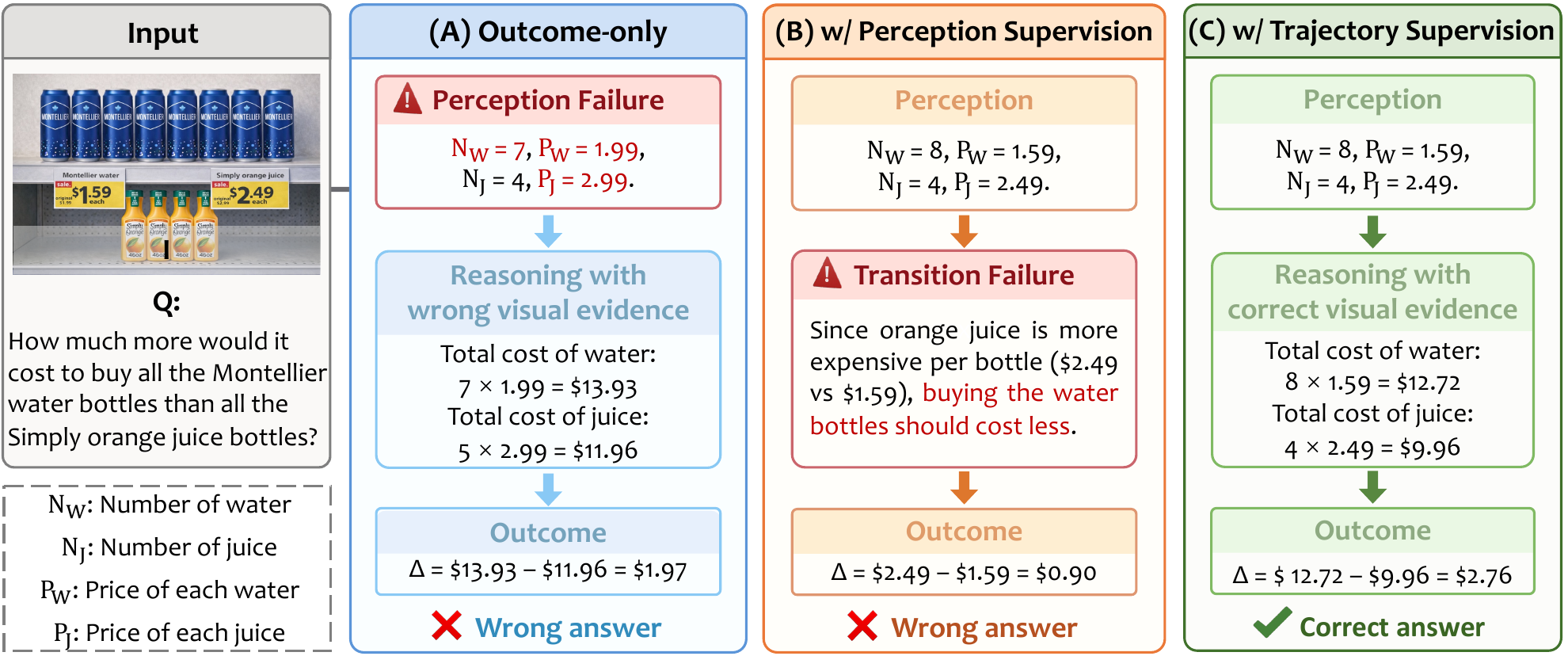}
    \caption{Comparison of different supervision strategies in multimodal RLVR. (A) Outcome-only supervision may lead to perception failures, where incorrect visual evidence is used for reasoning. (B) Adding perception supervision improves visual grounding but does not guarantee correct reasoning transitions. (C) Our trajectory supervision aligns the perception--reasoning process, enabling reasoning grounded in correct visual evidence and leading to correct answers.}
    \label{fig:intro}
\end{figure}

However, unlike purely linguistic reasoning in LLMs, multimodal reasoning requires jointly modeling \textbf{visual perception} and \textbf{coherent reasoning}. 
Recent works have identified perception as a key bottleneck in multimodal reasoning, as standard RLVR frameworks primarily optimize outcome-level correctness without enforcing visual grounding~\cite{PAPO, perception_r1}. 
As illustrated in Fig.~\ref{fig:intro}(A), without explicit supervision of the perceptual process, models may misinterpret visual content and produce reasoning chains based on incorrect perceptual signals.
To address this, recent studies attempt to strengthen the perceptual stage of multimodal reasoning through two main directions:
\begin{itemize}
 \item \textbf{Image Augmentation}: input-level transformations such as cropping, randomly masking, and noise injection that enhance the model's sensitivity to critical visual regions~\cite{NoisyRollout, PAPO};
 \item \textbf{Perception Reward}: auxiliary reward signals that explicitly guide the model toward task-relevant visual cues, often derived by comparing the model's outputs with those from a stronger model (\textit{e.g.}, Gemini-2.5 Pro)~\cite{perception_r1_2}.
\end{itemize}

However, these approaches mainly address \textbf{where} the model should look, rather than \textbf{how} visual evidence is incorporated into the reasoning process. 
Even when perception is accurate, existing RLVR frameworks do not explicitly enforce how perceived visual evidence should be translated into textual descriptions and utilized during reasoning, allowing models to produce answers without faithfully grounding on visual evidence, as illustrated in Fig. \ref{fig:intro}. (B).

In this paper, we revisit the optimization flow of multimodal RLVR and show that existing approaches either rely solely on outcome-level supervision or introduce supervision only at the perception stage, leaving the transition from perception to reasoning unconstrained (Fig. \ref{fig:intro}. (C)).
To address this limitation, we propose \textbf{T}rajectory-\textbf{G}uided \textbf{R}einforcement \textbf{L}earning (TGRL), which incorporates expert reasoning trajectories to align the perception--reasoning transition during RLVR training.
Expert trajectories provide explicit perception–reasoning paths that serve as structured supervision signals, enabling more effective credit assignment along the multimodal reasoning trajectory.

In summary, our contributions are threefold:
\begin{itemize}
\item We systematically analyze the optimization flow of multimodal RLVR and identify a critical bottleneck in the perception–reasoning transition.
\item We propose Trajectory-Guided Reinforcement Learning (TGRL), which introduces trajectory-level supervision to align the perception--reasoning transition during RLVR training.
\item Extensive experiments on various benchmarks demonstrate that TGRL consistently improves performance over existing RLVR baselines.
\end{itemize}
\section{Preliminaries}
\label{sec:preliminary}
In this section, we revisit Reinforcement Learning with Verifiable Rewards (RLVR) procedure and representative RLVR optimization strategies for multi-modal large language models (MLLMs). 

\subsection{Task Formulation}

\textbf{Reinforcement Learning with Verifiable Rewards (RLVR) for MLLMs}: RLVR enhances multi-modal large language models (MLLMs) by aligning model outputs with verifiable answers. Given a multi-modal input with ground truth from a batch of B samples, $ (\mathrm{q}^{b}, \mathrm{I}^{b}) \in \{\mathrm{q}^b, \mathrm{I}^b\}_{b=1}^{B}$, and $y^{b} \in \{ y^{b}\}_{b=1}^{B}$, where $\mathrm{I}^{b}$, $\mathrm{q}^{b}$, and $y^{b}$ denotes the image, question, and ground-truth respectively, the model $\bm{\pi}_{\theta}$ generates output $\bm{o}^{b}$ containing a reasoning process and a prediction $\hat{y}^{b}$. The prediction is enclosed in \texttt{\textbackslash boxed\{.\}} while reasoning is delimited by \texttt{<think>...</think>}, enabling automated verification against the ground truth answers.
RLVR employs a binary reward function $\mathbf{R}(\cdot)$ to determine whether the answer is correct by comparing the model prediction $\hat{y}^{b}$ with ground truth $y^{b}$.
The goal of RLVR is to maximize the reward function, formalized as:
\begin{equation}
\mathcal{J}_{\mathrm{RLVR}}(\theta) = 
\displaystyle \max_{\theta} 
\mathbb{E}_{\{(\mathrm{q}^b, \mathrm{I}^{b}), y^{b}) \}_{b=1}^{B}}
\mathbb{E}_{\bm{o}^{b} \sim \bm{\pi}_{\theta}(\cdot \mid \mathrm{q}^{b}, \mathrm{I}^{b})}[
\mathbf{R}(\hat{y}^{b}, y^{b})].
\end{equation}

\subsection{RLVR Algorithms}

\textbf{Group Relative Policy Optimization (GRPO).} As a widely adopted RLVR optimization strategy, GRPO stabilizes training by computing advantages within response groups~\cite{deepseekmath}. Concretely, given a batch of samples $\{(\mathrm{q}^b, \mathrm{I}^b), y^{b}\}_{b=1}^{B}$, the GRPO objective is:
\begin{equation}
\begin{aligned}
\mathcal{J}_{\text{GRPO}}&(\theta)  = \mathbb{E}_{\substack{\mathrm{q}^b, \mathrm{I}^b \sim \mathcal{D} \\ \bm{o}_i^b \sim \bm{\pi}_{\theta_{\text{old}}(\cdot)}}} \Bigg[ \frac{1}{G} \sum_{i=1}^G \frac{1}{|o_i^{b}|} \sum_{t=1}^{|o_i^{b}|} \min \bigg( \frac{\bm{\pi}_\theta(\bm{o}_{i,t}^{b}\mid \bm{o}^{b}_{i,<t},\mathrm{q}^{b},\mathrm{I}^{b})}{\bm{\pi}_{\theta_{\text{old}}}(\bm{o}^{b}_{i,t}\mid\bm{o}^{b}_{i,<t},\mathrm{q}^{b},\mathrm{I}^{b})} \hat{A}^{b}_{i}, \\
&\text{clip} \left( \frac{\bm{\pi}_\theta(\bm{o}^{b}_{i,t}\mid\bm{o}^{b}_{i,<t},\mathrm{q}^{b},\mathrm{I}^{b})}{\bm{\pi}_{\theta_{\text{old}}}(o^{b}_{i,t}\mid\bm{o}^{b}_{i,<t},\mathrm{q}^{b},\mathrm{I}^{b})}, 1 \pm \epsilon \right) \hat{A}^{b}_{i} \bigg) - \beta \cdot \mathbb{D}_{KL}(\bm{\pi}_\theta \| \bm{\pi}_{\text{ref}}) \Bigg],
\end{aligned}
\end{equation}
where $G$ is the rollout group size for each input, and the clip function restricts the importance ratio within $[1 - \epsilon, 1 + \epsilon]$. Moreover, the advantage is:
\begin{equation}
\begin{cases}
    \hat{A}^{b}_{i} = \dfrac{\mathbf{R}^{b}_i - \text{mean}(\mathbf{R}^{b})}{\text{std}(\mathbf{R}^{b})}, \\[10pt]
    \mathbf{R}^{b}_{i} = \mathbb{I}(\texttt{is\_equivalent}(\hat{y}^{b}, y^{b})),
\end{cases}
\end{equation}
where $\mathbb{I}(\cdot)$ is the indicator function, $\hat{y}^{b}$ is the extracted predictions from the model output $\bm{o}^{b}_{i}$, and $\texttt{is\_equivalent}(\cdot)$ compares the extracted prediction $\hat{y}^{b}$ with the ground truth $y^{b}$.

\noindent \textbf{Decoupled Clip and Dynamic Sampling Policy Optimization (DAPO)}.
DAPO~\cite{dapo} improves upon GRPO by removing KL regularization and introducing clip-higher, dynamic sampling, and token-level loss from the GRPO loss, achieving new state-of-the-art performance.
Specifically, the DAPO objective for the batch of samples $\{(\mathrm{I}^b, \mathrm{q}^b), y^{b}\}_{b=1}^{B}$ can be formalized as:
\begin{equation}
\begin{aligned}
\mathcal{J}_{\text{DAPO}}&(\theta) = \mathbb{E}_{\substack{\mathrm{q}^{b}, \mathrm{I}^{b} \sim \mathcal{D} \\ \bm{o}^{b}_i \sim \bm{\pi}_{\theta_{\text{old}}(\cdot)}}} \Bigg[ \frac{1}{\sum |\bm{o}_i^{b}|} \sum_{i=1}^{G} \sum_{t=1}^{|\bm{o}^{b}_i|} \min \bigg( \frac{\bm{\pi}_\theta(\bm{o}^{b}_{i,t}\mid\bm{o}^{b}_{i,<t},\mathrm{q}^{b},\mathrm{I}^{b})}{\bm{\pi}_{\theta_{\text{old}}}(o^{b}_{i,t}\mid\bm{o}^{b}_{i,<t},\mathrm{q}^{b},\mathrm{I}^{b})} \hat{A}^{b}_{i}, \\
& \text{clip} \left( \frac{\bm{\pi}_\theta(\bm{o}^{b}_{i,t}\mid\bm{o}^{b}_{i,<t},\mathrm{q}^{b},\mathrm{I}^{b})}{\bm{\pi}_{\theta_{\text{old}}}(\bm{o}_{i,t}\mid\bm{o}^{b}_{i,<t},\mathrm{q}^{b},\mathrm{I}^{b})}, 1 - \epsilon_{\text{low}}, 1 + \epsilon_{\text{high}} \right) \hat{A}^{b}_{i} \bigg) \Bigg],
\end{aligned}
\end{equation}
where $\epsilon_{\text{low}}$ and $\epsilon_{\text{high}}$ decouple the clipping thresholds for negative and positive advantages, respectively. In this work, we apply our token-reweighting strategy to both GRPO and DAPO, demonstrating its general applicability across various RLVR optimization strategies.

\section{Revisiting Optimization Flow in Multimodal RLVR}
\label{sec:analysis}

\subsection{The Multimodal Reasoning Pipeline}
To understand the limitations of current paradigms, we formalize the generation process of an MLLM $\bm{\pi}_{\theta}$. 
Given a question $\mathrm{q}^{b}$ and an image $\mathrm{I}^b$ from a batch $\{(\mathrm{q}^b, \mathrm{I}^b), y^{b}\}_{b=1}^{B}$, the generation process of a MLLM $\bm{\pi}_{\theta}$ can be decomposed into:
\begin{equation}
(\mathrm{q}^b, \mathrm{I}^b)
\xrightarrow{\bm{\pi}_\theta} \mathrm{p}^b
\xrightarrow{\bm{\pi}_\theta} \mathrm{r}^b
\xrightarrow{g(\cdot)} \hat{y}^b,
\end{equation}
where $\mathrm{p}^{b}$ represents responses describing the image $\mathrm{I}^b$, and $\mathrm{r}^{b}$ denotes the subsequent reasoning responses conditioned on $(\mathrm{p}^b, \mathrm{q}^b)$. 
Together, they form the complete response $\bm{o}^b = \{\mathrm{p}^b, \mathrm{r}^b\}$, from which the final prediction $\hat{y}^b$ is derived as $\hat{y}^b = g(\mathrm{r}^b)$ using an extraction function $g(\cdot)$.
Under this formulation, the distribution of generating the trajectory $\bm{o}^b$ can be factorized as:
\begin{equation}
    \bm{\pi}_{\theta}(\bm{o}^{b}|\mathrm{q}^{b},\mathrm{I}^{b}) =  \underbrace{\colorbox{percolor!9}{$\displaystyle \bm{\pi}_{\theta}(\mathrm{p}^{b}|\mathrm{q}^{b},\mathrm{I}^{b})$}}_{\text{Perception}}
    \cdot \underbrace{\colorbox{reacolor!9}{$\displaystyle \bm{\pi}_{\theta}(\mathrm{r}^{b}|\mathrm{p}^{b}, \mathrm{q}^{b})$}}_{\text{Reasoning}}.
\end{equation}
This factorization exposes a two-stage dependency:
\begin{itemize}
    \item Perception modeling: $\bm{\pi}_{\theta}(\mathrm{p}^{b}|\mathrm{q}^{b},\mathrm{I}^{b})$
    \item Perception to reasoning transition: $\bm{\pi}_{\theta}(\mathrm{r}^{b}|\mathrm{p}^{b}, \mathrm{q}^{b})$
\end{itemize}

\subsection{Bottlenecks in Current Optimization Flows}
The essential difference among existing approaches lies in which components of this pipeline are optimized, and how reward is assigned along $(\mathrm{p} \rightarrow \mathrm{r} \rightarrow \hat{y})$ during RLVR. 
We show how existing RLVR strategies distribute reward signals along the multimodal reasoning trajectory.

\noindent \ding{182} \textbf{Outcome-Level Supervision.}
Standard outcome-level RLVR optimizes the model solely based on the correctness of the final prediction:
\begin{equation}
\mathcal{J}(\theta) 
= \max_{\theta}\; 
\mathbb{E}_{\bm{o}^{b} \sim \bm{\pi}_\theta(\cdot \mid \mathrm{q}^{b}, \mathrm{I}^{b})}
\big[\mathbf{R}(\hat{y}^{b}, y^{b})\big],
\end{equation}
where $\bm{o}^b = (\mathrm{p}^b, \mathrm{r}^b)$ is the generated trajectory,
$\hat{y}^{b} = g(\mathrm{r}^b)$ is the extracted prediction, and $\mathbf{R}$ is a binary reward.
Under the simple REINFORCE objective \cite{reinforce}, the gradient is derived as:
\begin{equation}
\begin{aligned}
\nabla_\theta & \mathcal{J}(\theta) 
= \mathbb{E}_{\bm{o}^b \sim \bm{\pi}_\theta(\cdot \mid \mathrm{q}^{b}, \mathrm{I}^{b})}\Big[\mathbf{R}(\hat{y}^{b}, y^{b}) \cdot \nabla_\theta \log \bm{\pi}_\theta(\bm{o}^b \mid \mathrm{q}^{b}, \mathrm{I}^{b})\Big] \\
= & \mathbb{E}_{\bm{o}^b \sim \bm{\pi}_\theta(\cdot \mid \mathrm{q}^{b}, \mathrm{I}^{b})}\Big[\mathbf{R}(\hat{y}^{b}, y^{b}) \cdot  \underbrace{\colorbox{cyan!5}{$\displaystyle \Big( \nabla_\theta \log \bm{\pi}_\theta( \mathrm{r}^{b} | \mathrm{p}^{b}, \mathrm{q}^{b}) + \nabla_\theta \log \bm{\pi}_\theta( \mathrm{p}^{b} | \mathrm{q}^{b}, \mathrm{I}^{b}) \Big)$}}_{\text{Outcome gradient on full trajectory}} \Big]
\end{aligned}
\end{equation}
\textbf{Limitations.} 
Although the gradient propagates through both perception and reasoning components, the credit assignment signal is entirely determined by the final correctness.
As a result: \textbf{(1)} The perception module is not explicitly grounded in verifiable visual evidence.
\textbf{(2)} The reasoning module is not required to faithfully utilize the perceived facts.
This outcome-level supervision allows the model to obtain correct answers via spurious correlations or implicit language priors, rather than through consistent perception---reasoning alignment.

\noindent \ding{183} \textbf{Perception-Level Supervision.}
To provide denser signals for the perceptual stage, perceptual reward $\mathbf{R}_{\text{per}}$ is introduced to compare the perceived representation $\mathrm{p}^b$ with a supervision target $\mathcal{G}^b$ (\textit{e.g.}, grounding annotations):
\begin{equation}
\mathcal{J}(\theta)
=
\max_{\theta}\;
\mathbb{E}_{\bm{o}^b \sim \bm{\pi}_\theta(\cdot \mid \mathrm{q}^b, \mathrm{I}^b)}
\Big[
\underbrace{\mathbf{R}(\hat{y}^b, y^b)}_{\text{Outcome}}
+
\lambda\, \underbrace{\mathbf{R}_{\text{per}}(\mathrm{p}^b, \mathcal{G}^b)}_{\text{Perception}}
\Big].
\label{eq:per_obj}
\end{equation}
Its gradient becomes:
\begin{equation}
\begin{aligned}
\nabla_\theta \mathcal{J}(\theta) = \mathbb{E}_{\bm{o}^b \sim \bm{\pi}_\theta(\cdot \mid \mathrm{q}^b, \mathrm{I}^b)} \Big[ & \underbrace{\mathbf{R}(\hat{y}^b,{y}^b) \cdot \nabla\theta \log \bm{\pi}_\theta(\bm{o}^b \mid \mathrm{q}^{b}, \mathrm{I}^{b})}_{\text{Outcome Gradient}} \\ + & \underbrace{\colorbox{percolor!9}{$\displaystyle \lambda \cdot \mathbf{R}_{\text{per}}(p^b,\mathcal G^b) \cdot \nabla_\theta \log \bm{\pi}_\theta(\mathrm{p}^b \mid \mathrm{q}^{b}, \mathrm{I}^{b})
$}}_{\text{Perception Gradient}} \Big],
\end{aligned}
\end{equation}
\textbf{Limitations.} 
Perception-level supervision improves what is observed, but leaves the perception---reasoning coupling unconstrained.
Although the auxiliary reward explicitly supervises the perception module, it does not constrain how the perceived evidence $\mathrm{p}^b$ is subsequently utilized in reasoning.
In particular, the transition distribution $\bm{\pi}_\theta(\mathrm{r}^b \mid \mathrm{p}^b, \mathrm{q}^b)$ remains optimized solely through outcome-level feedback.
Therefore, even with accurate perceptual grounding, the reasoning process is not required to faithfully ground on $\mathrm{p}^b$.

\subsection{The Necessity of Trajectory-Level Supervision}
The fundamental weakness of the above approaches lies in their structural decoupling of perception and reasoning.
While outcome-level supervision relies on final rewards and perception-level supervision constrains only intermediate representations,
neither explicitly regulates the transition from perception to reasoning.
To explicitly supervise this transition, we introduce trajectory-level supervision that aligns the perception---reasoning path ($\mathrm{p} \rightarrow \mathrm{r}$) with expert trajectories.

Let $\bm{o}_\phi^b=(\mathrm{p}_\phi^b, \mathrm{r}_\phi^b)$ denote an expert-generated trajectory sampled from  $\pi_\phi(\cdot\mid \mathrm{q}^b,\mathrm{I}^b)$.
We optimize the following objective:
\begin{equation}
\mathcal{J}(\theta)
= \max_{\theta} \mathbb{E}_{
\substack{_{\bm{o}_{\theta}^b \sim \bm{\pi}_\theta(\cdot \mid \mathrm{q}^b, \mathrm{I}^b)} \\
{\bm{o}_{\phi}^b \sim \bm{\pi}_\phi(\cdot \mid \mathrm{q}^b, \mathrm{I}^b)}}}
\Big[ \underbrace{\mathbf{R}(\hat{y}_{\theta}^{b}, y^{b})}_{\text{Outcome}} + 
\underbrace{\bm{f}(\bm{o}^{b}_{\phi};\theta)
\;\mathbf{R}(\hat{y}_{\phi}^{b},y^b)}_{\text{Expert Trajectory}}
\Big],
\label{eq:tgrl_obj}
\end{equation}
where $\hat{y}_\theta^b=g(r_\theta^b)$, $\hat{y}_\phi^b=g(r_\phi^b)$.
The token-level weighting function $\bm{f}(\bm{o}^{b}_{\phi};\theta)$ modulates the contribution of each token in expert trajectories, encouraging the student policy to align with structurally coherent perception–reasoning paths. 
Since $\bm{o}_\phi^b=\{\mathrm{p}_\phi^b,\mathrm{r}_\phi^b \}$
combines perception and reasoning tokens, we decompose $\bm{f}(\bm{o}^{b}_{\phi};\theta)$ according to  $\mathrm{p}_\phi^b$ and $\mathrm{r}_\phi^b$ as $\bm{f}(\mathrm{p}_\phi^b;\theta)$ and $\bm{f}(\mathrm{r}_\phi^b;\theta)$, respectively.
Under the policy gradient framework, the gradient decomposes as: 
\begin{equation}
\begin{aligned}
\nabla_\theta & \mathcal{J}(\theta)
=
\mathbb{E}_{
\substack{_{\bm{o}_{\theta}^b \sim \bm{\pi}_\theta(\cdot \mid \mathrm{q}^b, \mathrm{I}^b)} \\
{\bm{o}_{\phi}^b \sim \bm{\pi}_\phi(\cdot \mid \mathrm{q}^b, \mathrm{I}^b)}}}\Big[
\mathbf{R}(\hat{y}_{\theta}^{b}, y^{b})
\cdot \nabla_\theta \log \bm{\pi}_\theta(\bm{o}_{\theta}^b\mid \mathrm{q}^{b}, \mathrm{I}^{b})
+ \mathbf{R}(\hat{y}_{\phi}^{b}, y^{b})
\cdot \\
&
\Big( \underbrace{
\colorbox{percolor!9}{$\displaystyle
\bm{f}(\mathrm{p}^{b}_{\phi};\theta) \cdot \nabla_\theta \log \bm{\pi}_{\theta}(\mathrm{p}_{\phi}^{b}\mid \mathrm{q}^{b}, \mathrm{I}^{b})$}}_{\text{Perception-aligned Gradient}} 
+ \underbrace{\colorbox{reacolor!9}{$\displaystyle\bm{f}(\mathrm{r}^{b}_{\phi};\theta) \cdot \nabla_\theta \log \bm{\pi}_{\theta}(\mathrm{r}_{\phi}^{b}\mid \mathrm{p}_{\phi}^{b},\mathrm{q}^{b}) $}}_{\text{Reasoning-aligned Gradient}}
\Big)
\Big].
\end{aligned}
\end{equation}
Unlike outcome-level or perception-level supervision, trajectory-level alignment directly constrains the transition from perception to reasoning, enabling structured credit assignment across the entire multimodal reasoning pipeline.
\section{Trajectory-Guided Reinforcement Learning}
\label{sec:approach}
\begin{figure}
    \centering
    \includegraphics[width=\linewidth]{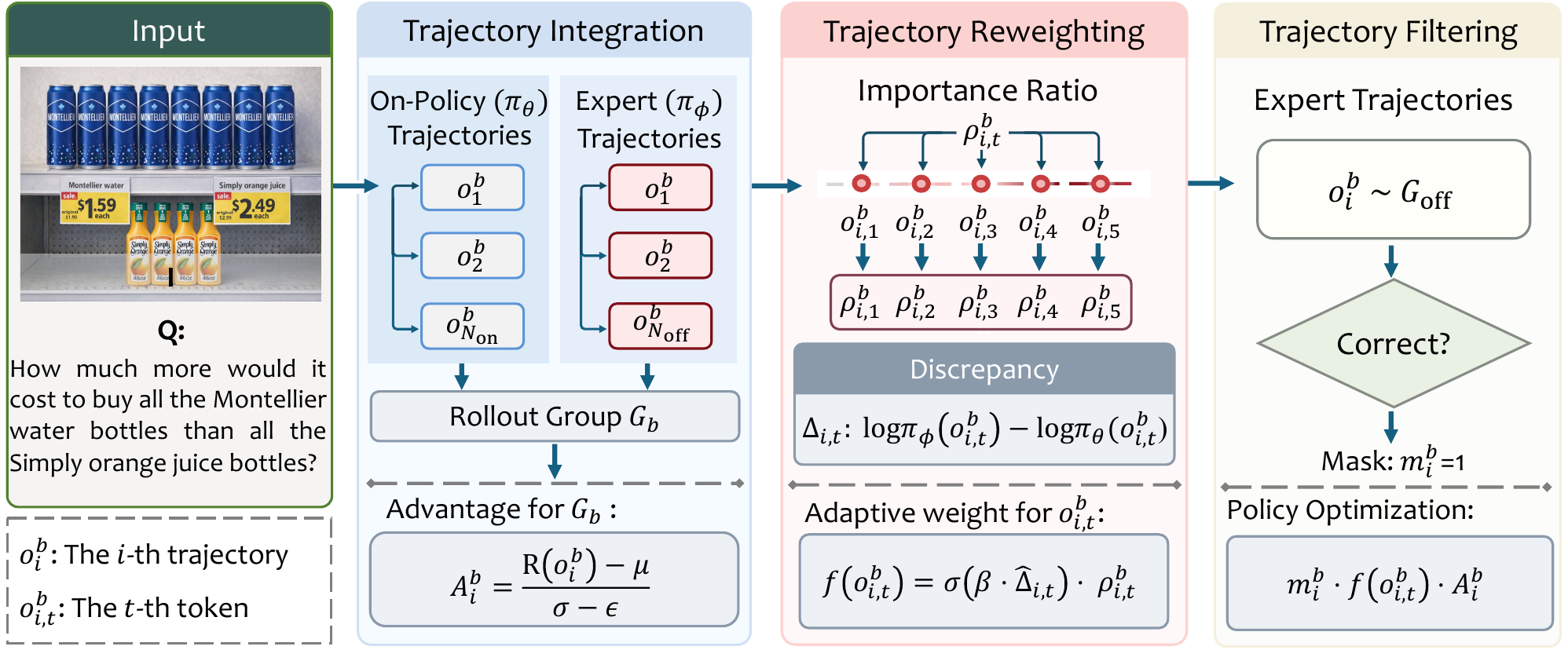}
    \caption{Overview of \textbf{T}rajectory \textbf{G}uided \textbf{R}einforcement \textbf{L}earning (TGRL). Given an image-question pair, we construct a rollout group by integrating on-policy and expert trajectories. Expert trajectories are then reweighted at the token level and filtered by correctness, enabling trajectory-level alignment within RLVR framework.}
    \label{fig:method}
\end{figure}

Motivated by the gradient analysis, we propose a \textbf{T}rajectory \textbf{G}uided \textbf{R}einforcement \textbf{L}earning (TGRL)  that implements such trajectory-level alignment into the RLVR framework. An overview of TGRL is illustrated in Fig.~\ref{fig:method}.

\subsection{Trajectory Integration}
As discussed in Section~\ref{sec:analysis}, trajectory-level supervision requires aligning current policy with expert trajectories.
However, directly fine-tuning on expert trajectories suffers from distribution mismatch \cite{onpolicydistillation, luffy}, leading to error accumulation in reasoning trajectories, as early deviations push the model toward states outside its training distribution.
Instead of directly fine-tuning, we incorporate expert trajectories into a reinforcement learning framework, allowing expert rollouts to serve as guidance signals while preserving on-policy exploration.

\noindent $\bullet$ \textbf{Group Construction.}
Given a question $\mathrm{q}^b$ with image $\mathrm{I}^b$, we construct the rollout group consisting of both on-policy and expert-generated trajectories:
\begin{equation}
\begin{aligned}
\mathcal G_{\text{on}}^b & =\{\bm{o}_i^b \sim \bm{\pi}_{\theta_{\text{old}}}(\cdot \mid \mathrm{q}^b, \mathrm{I}^b)\}_{i=1}^{N_{\text{on}}}, \\
\mathcal G_{\text{off}}^b &=\{\bm{o}_i^b \sim \bm{\pi}_{\phi}(\cdot \mid \mathrm{q}^b, \mathrm{I}^b)\}_{i=1}^{N_{\text{off}}}.
\end{aligned}
\end{equation}
During optimization, trajectories are sampled from the previous policy $\bm{\pi}_{\theta_{\text{old}}}$, following standard PPO-style updates. Then, the combined rollout group is: 
\begin{equation}
    \mathcal G^b = \mathcal G_{\text{on}}^b \cup \mathcal G_{\text{off}}^b.
\end{equation}
Note that $\mathcal{G}_{\text{off}}^b$ typically contains a small number of expert trajectories, ensuring that the optimization remains predominantly on-policy.

\noindent $\bullet$ \textbf{Group-Based Advantage Computation.}
For each trajectory $\bm{o}_i^b \in \mathcal G^b$, we compute its reward
$ \mathbf{R}(\bm{o}_i^b) = \mathbf{R}(\hat y_i^b, y^b)$. 
To enable relative comparison within the group, we normalize the rewards with the mean $\mu_{\mathcal{G}^b}$ and standard deviation $\sigma_{\mathcal{G}^b}$:
\begin{equation}
    \mu_{\mathcal{G}^b} = \frac{1}{|\mathcal G^b|} \sum_{\bm{o}_{i}^{b} \in \mathcal G^b} \mathbf{R}(\bm{o}_{i}^{b}), \quad
    \sigma_{\mathcal{G}^b} = \mathrm{Std}\big(\{\mathbf{R}(\bm{o}_{i}^{b})\}_{\bm{o}_{i}^{b}\in\mathcal G^b}\big).
\end{equation}
Then, the normalized advantage is defined as:
\begin{equation}
\label{equ:normalized_advantage}
    \bar{A}_i^b = \frac{ \mathbf{R}(\bm{o}_{i}^{b}) - \mu_{\mathcal{G}^b}}{\sigma_{\mathcal {G}^b} + \epsilon}.
\end{equation}
Since advantages are normalized across the joint rollout group, trajectories with relatively higher rewards receive proportionally larger policy updates. In practice, expert trajectories often exhibit higher relative advantages during early training, thus contributing more strongly to the updates. As training progresses, the relative advantage of on-policy rollouts increases, gradually transitioning from expert guidance to on-policy exploration.

\subsection{Trajectory Reweighting}
To instantiate the weighting function introduced in Section~\ref{sec:analysis}, we design a two-stage reweighting mechanism.
We first perform importance ratio correction to account for off-policy expert trajectories.
Moreover, we introduce adaptive token-level reweighting to emphasize positions where the policy model deviates most from the expert.

\noindent $\bullet$ \textbf{Importance Ratio Correction.}
Expert trajectories are sampled from a stronger model $\bm{\pi}_\phi$ rather than the previous policy $\bm{\pi}_{\theta_{\text{old}}}$. 
Consequently, the standard on-policy importance sampling ratio $\bm{\pi}_{\theta} / \bm{\pi}_{\theta_{\text{old}}}$ is not applicable to these trajectories \cite{deepseekmath, ppo}. 
To account for this behavior mismatch, we define the importance ratio with respect to the expert policy $\bm{o}_{i}^b \sim \mathcal{G}_{\text{off}}^{b}$ as:
\begin{equation}
\rho_{\theta,i,t}^b = \frac{\bm{\pi}_\theta(\bm{o}_{i,t}^b \mid \bm{o}_{i,<t}^b,\mathrm{q}^b, \mathrm{I}^b)}{\bm{\pi}_\phi(\bm{o}_{i,t}^b \mid \bm{o}_{i,<t}^{b}, \mathrm{q}^b, \mathrm{I}^b)}.
\end{equation}
This ratio performs token-level off-policy correction, compensating for the distribution shift between expert-generated trajectories and the current policy.

\noindent $\bullet$ \textbf{Adaptive Token Reweighting.}
While the modified importance sampling ratio corrects for distribution mismatch, not all tokens in expert trajectories provide equally informative guidance. 
Intuitively, tokens where the current policy significantly underestimates the expert probability indicate larger discrepancies and therefore contain stronger corrective signals.
To capture this effect, we introduce a token-level reweighting coefficient. For an expert trajectory $\bm{o}_{i}^b=(\bm{o}_{i,1}^b,\dots,\bm{o}_{i,T}^b) \sim \mathcal{G}^{b}$, we define the discrepancy at step $t$ as:
\begin{equation}
\Delta_{i,t}^b = \log \bm{\pi}_\phi(\bm{o}_{i,t}^b \mid \bm{o}_{i,<t}^b, \mathrm{q}^b, \mathrm{I}^b) - \log \bm{\pi}_\theta(\bm{o}_{i,t}^b \mid \bm{o}_{i,<t}^b, \mathrm{q}^b, \mathrm{I}^b),
\end{equation}
which measures how much the current policy underestimates the expert token under the same prefix.
To stabilize optimization, we normalize the discrepancy within each trajectory:
\begin{equation}
\tilde{\Delta}_{i,t}^b = \frac{\Delta_{i,t}^b - \mu_{\Delta i}^{b}}{\sigma_{\Delta i}^{b} + \epsilon}.
\end{equation}
We then compute a bounded token weight as:
\begin{equation}
w_{i,t}^b = \sigma(\beta \tilde{\Delta}_{i,t}^b),
\end{equation}
where $\sigma(\cdot)$ denotes the sigmoid function and $\beta$ controls the sensitivity to discrepancy magnitude.
Finally, we define the modulated token-level importance weight, the weight at step $t$ is:
\begin{equation}
\bm{f}(\bm{o}_{i,t}^b; \theta) = \rho_{\theta,i,t}^b \cdot w_{i,t}^b,
\end{equation}
This formulation preserves token-level off-policy correction, while adaptively scaling gradient magnitudes according to the discrepancy between the current and expert policies.

\subsection{Trajectory Filtering}
Although expert-generated trajectories generally exhibit higher reasoning quality, they do not guarantee correct final predictions.
Expert trajectories with incorrect final predictions represent failed reasoning paths. Supervising the model with these samples introduces gradient noise that contradicts the outcome-based reward, thereby destabilizing the optimization process and hindering policy improvement.
In contrast, negative samples are more reliably obtained through on-policy exploration, which better reflects the model’s current uncertainty and failure modes~\cite{deepseek_v3_2}.
Therefore, we restrict expert supervision to trajectories that yield correct outcomes.

\noindent $\bullet$ \textbf{Positive Trajectory Filtering.}
We apply a correctness-based filtering mechanism
that selectively incorporates only successful expert trajectories.
For an expert-generated trajectory $\bm{o}_{\phi}^b$, we define a binary mask:
\begin{equation}
m_{\phi}^b =
\begin{cases}
1, & \mathbf{R}(\hat{y}_{\phi}^b, y^b) = 1, \\
0, & \mathbf{R}(\hat{y}_{\phi}^b, y^b) = 0,
\end{cases}
\end{equation}
where $\hat{y}_\phi^b = g(\mathrm{r}_\phi^b)$.
The masked expert contribution is then incorporated into the optimization
only when $m_{\phi}^b = 1$,
ensuring that trajectory-level alignment is guided by correct reasoning paths.

\subsection{Training Objective}
To incorporate expert-generated trajectories into the RLVR framework, TGRL modifies the optimization in three aspects: 
\textbf{(1)} the rollout distribution,
\textbf{(2)} the advantage normalization scope,
and \textbf{(3)} the token-level importance weighting.
We incorporate trajectory integration, adaptive reweighting, and positive filtering into GRPO~\cite{deepseekmath} and DAPO~\cite{dapo} by replacing the standard importance ratio with a unified token-level coefficient.
For each trajectory $\bm{o}_i^b \in \mathcal{G}^b$, we define the unified token-level coefficient as:
\begin{equation}
\tilde{r}_{i,t}^b =
\begin{cases}
\dfrac{
\bm{\pi}_\theta(\bm{o}_{i,t}^b \mid \bm{o}_{i,<t}^b, \mathrm{q}^b, \mathrm{I}^b)
}{
\bm{\pi}_{\theta_{\text{old}}}(\bm{o}_{i,t}^b \mid \bm{o}_{i,<t}^b, \mathrm{q}^b, \mathrm{I}^b)
},
& \bm{o}_i^b \in \mathcal{G}_{\text{on}}^b,
\\[10pt]
m_\phi^b \cdot
\bm{f}(\bm{o}_{\phi,t}^b; \theta),
& \bm{o}_i^b \in \mathcal{G}_{\text{off}}^b,
\end{cases}
\end{equation}
where $\bm{f}(\bm{o}_{\phi,t}^b; \theta) = \rho_{\theta,t}^b \cdot w_t^b$
combines importance ratio correction and token-level reweighting strategy, and $m_\phi^b$ filters out incorrect expert trajectories.

\noindent $\bullet$ \textbf{TGRL-GRPO.} 
We extend GRPO by performing optimization over the joint rollout group $\mathcal{G}^b = \mathcal{G}_{\text{on}}^b \cup \mathcal{G}_{\text{off}}^b$. 
By normalizing advantages across this unified group as in Equation~\ref{equ:normalized_advantage} and replacing the standard importance ratio with the coefficient $\tilde r_{i,t}^b$, we obtain the TGRL-GRPO objective as:
\begin{equation}
\begin{aligned}
\mathcal{J}_{\text{TGRL-GRPO}}(\theta) = &\mathbb{E}_{\substack{ (\mathrm{q}^b,\mathrm{I}^b)\sim\mathcal{D} \\ \bm{o}_i^b \sim \mathcal{G}^b }} \Bigg[ \frac{1}{G}\sum_{i=1}^{G} \frac{1}{|\bm{o}_i^b|} \sum_{t=1}^{|\bm{o}_i^b|} \min \Big( \tilde r_{i,t}^b \bar A_{i}^b, \\ 
& \text{clip}(\tilde r_{i,t}^b, 1-\epsilon, 1+\epsilon) \bar A_{i}^b \Big) - \beta \mathbb{D}_{KL}(\bm{\pi}_\theta \| \bm{\pi}_{\text{ref}}) \Bigg].
\end{aligned}
\end{equation}

\noindent $\bullet$ \textbf{TGRL-DAPO.}
Similarly, we extend DAPO under the same trajectory-guided framework.
The resulting TGRL-DAPO objective is:
\begin{equation}
\begin{aligned}
\mathcal{J}_{\text{TGRL-DAPO}}(\theta) = &\mathbb{E}_{\substack{ (\mathrm{q}^b,\mathrm{I}^b)\sim \mathcal{D} \\ \bm{o}_i^b \sim \mathcal{G}^b }} \Bigg[ \frac{1}{\sum |\bm{o}_i^b|} \sum_{i=1}^{G} \sum_{t=1}^{|\bm{o}_i^b|} \min \Big( \tilde r_{i,t}^b \bar A_{i}^b, \\
& \text{clip}(\tilde r_{i,t}^b, 1-\epsilon_{\text{low}}, 1+\epsilon_{\text{high}}) \bar A_{i}^b \Big) \Bigg].
\end{aligned}
\end{equation}

\noindent \textbf{Discussion.}
TGRL incorporates expert trajectories into RLVR by modifying the rollout distribution, advantage normalization, and token-level importance weighting. 
The resulting objectives preserve the underlying gradient structure of GRPO-style RLVR objectives while enabling trajectory-level alignment, achieving a principled balance between expert guidance and on-policy exploration.
\section{Experiments}
\label{sec:expriment}
In this section, we validate the effectiveness of \textbf{T}rajectory-\textbf{G}uided \textbf{R}einforcement \textbf{L}earning (TGRL) by answering the following \textbf{R}esearch \textbf{Q}uestions (RQs):

\noindent \ding{182} How do different components of TGRL affect overall performance?

\noindent \ding{183} How sensitive is TGRL to the hyperparameter $\beta$?

\noindent \ding{184} How does TGRL compare with state-of-the-art approaches?

\subsection{Experimental Settings}

\noindent \textbf{Training Data.} 
We conduct training on two multimodal reasoning datasets: Geo3K~\cite{geo3K}, which contains 2.1K geometry reasoning samples, and ViRL39K~\cite{vl_rethinker}, a large-scale dataset with 39K multimodal reasoning instances.

\noindent \textbf{Implementation Details.}
All experiments are implemented using the EasyR1 training framework\footnote{\url{https://github.com/hiyouga/EasyR1}}. 
We adopt Qwen2.5-VL-3B and Qwen2.5-VL-7B as the student policy $\pi_{\theta}$ and Qwen2.5-VL-32B as the expert policy $\pi_{\phi}$ for trajectory generation. 
For Geo3K training, the rollout group size is set to $8$, consisting of $N_{\text{on}}=7$ on-policy trajectories and $N_{\text{off}}=1$ expert trajectory.
For ViRL39K training, the group size is $5$, with $N_{\text{on}}=4$ and $N_{\text{off}}=1$.
Models are trained for $15$ epochs on Geo3K and $2$ epochs on ViRL39K.
Unless otherwise specified, all other hyperparameters follow the default settings.

\noindent \textbf{Evaluation Benchmarks.}
We evaluate models on both multimodal reasoning and perception benchmarks.
\textit{Reasoning benchmarks} include MathVista~\cite{mathvista}, WeMath~\cite{wemath}, MathVision~\cite{mathvision}, and MathVerse~\cite{mathverse}.
\textit{Perception benchmarks} include HallusionBench~\cite{hallusionbench}.

\subsection{Ablation Studies}

\begin{table*}[t]
\caption{
Component ablation of TGRL on Qwen2.5-VL-7B under GRPO and DAPO.
Removing trajectory reweighting or filtering consistently degrades performance, demonstrating the importance of properly utilizing expert trajectories. We set $\beta=5$ for GRPO and $\beta=14$ for DAPO.
}
\label{tab:ablation_components}
\centering
\small
\setlength{\tabcolsep}{2pt}
\begin{tabular}{l c c c c c c}
\toprule
Method & WeMath & MathVista & MathVerse & MathVision & HalluBench & Avg. \\
\midrule
\textbf{TGRL-GRPO} & 70.29 & 71.50 & 51.08 & 27.41 & 72.02 & 58.46\\
w/o Filter & 68.74 & 70.80 & 50.65 & 26.68 & 71.04 & 57.50 \\
w/o Reweight & 67.58 & 70.40 & 50.14 & 27.11 & 69.90 & 57.03 \\
w/o Expert (GRPO) & 67.40 & 70.50 & 50.80 & 27.30 & 69.80 & 57.16 \\
\midrule
\textbf{TGRL-DAPO} & 71.05 & 71.40 & 55.21 & 28.61 & 71.29 & 59.51\\
w/o Filter & 69.80 & 70.20 & 51.74& 27.57 & 70.35 & 57.93 \\
w/o Reweight & 67.93 & 70.90 & 50.93 & 27.20 & 70.56 & 57.50 \\
w/o Expert (DAPO) & 69.30 & 70.30 & 50.60 & 26.50 & 67.90 & 56.92 \\
\bottomrule
\end{tabular}
\end{table*}

\begin{table*}[t]
\caption{
Sensitivity analysis of the token reweighting coefficient $\beta$ on the Qwen2.5-VL-7B under GRPO and DAPO.
Baseline denotes vanilla GRPO/DAPO training without trajectories. Best results within each block are highlighted in \textbf{bold}.
}
\label{tab:beta}
\centering
\small
\setlength{\tabcolsep}{3.5pt}
\begin{tabular}{l c c c c c c}
\toprule
Method & WeMath & MathVista & MathVerse & MathVision & HalluBench & Avg. \\
\midrule
GRPO & 67.40 & 70.50 & 50.80 & 27.30 & 69.80 & 57.16 \\
$\beta=1$  &68.70&70.60&51.03&27.68&71.40&57.88\\
$\beta=4$  &69.37&70.80&\textbf{51.14}&27.60&71.71&58.12\\
$\mathbf{\beta=5}$ &\textbf{70.29}&\textbf{71.50}&51.08&27.41&\textbf{72.02}&\textbf{58.46}\\
$\beta=7$  &68.10&70.54&50.53&27.50&70.14&57.36\\
$\beta=10$ &68.36&70.24&50.81&\textbf{27.83}&69.62&57.37\\
$\beta=13$ &68.41&70.37&50.53&27.43&69.93&57.33\\
$\beta=15$ &68.74&69.80&50.78&27.77&71.29&57.68\\
\midrule
DAPO & 69.30 & 70.30 & 50.60 & 26.50 & 67.90 & 56.92 \\
$\beta=1$  &70.50&70.20&51.34&26.81&69.82&57.73\\
$\beta=4$  &70.20&70.90&50.93&26.67&70.56&57.85\\
$\beta=7$  &70.97&70.90&51.86&26.24&69.72&57.94\\
$\beta=10$ &69.22&\textbf{71.50}&51.84&\textbf{30.26}&70.24&58.61\\
$\beta=13$ &69.98&70.47&54.10&27.84&70.24&58.53\\
$\mathbf{\beta=14}$ &\textbf{71.05}&71.40&\textbf{55.21}&28.61&\textbf{71.29}&\textbf{59.51}\\
$\beta=15$ &70.90&71.10&51.40&25.67&70.35&57.88\\
\bottomrule
\end{tabular}
\end{table*}

\noindent $\bullet$ \textbf{Component Analysis (RQ1).}
Table~\ref{tab:ablation_components} shows the contribution of each component in TGRL.
Interestingly, directly introducing expert trajectories does not consistently improve performance, highlighting the importance of properly utilizing expert supervision.
Under the GRPO framework, simply introducing expert trajectories brings only marginal gains over the baseline, and removing the reweighting module even leads to performance degradation.
This observation suggests that expert trajectories alone may introduce noisy supervision signals.

The proposed trajectory reweighting and filtering mechanisms address this issue by emphasizing informative tokens and discarding incorrect expert trajectories.
By combining these components, TGRL achieves the best overall performance.
Specifically, TGRL improves the average score $57.16$ $\to$ $58.46$ under GRPO and $56.92$ $\to$ $59.51$ under DAPO, demonstrating the effectiveness of properly leveraging expert trajectories.

\noindent $\bullet$ \textbf{Hyperparameter Analysis (RQ2).}
We analyze the sensitivity of the token-reweighting coefficient $\beta$, which controls the strength of discrepancy-based token weighting. 
Table~\ref{tab:beta} reports the performance under different $\beta$ values.

Moderate values of $\beta$ generally yield the best performance. 
Specifically, the optimal value is $\beta=5$ under GRPO and $\beta=14$ under DAPO.
This difference arises from their optimization structures: GRPO performs trajectory-level averaging with group-normalized advantages, while DAPO averages gradients across all tokens, resulting in more uniformly distributed gradient signals.
Consequently, DAPO requires a stronger token-weighting coefficient to highlight informative tokens.
Overall, TGRL remains robust across a wide range of $\beta$ values and consistently improves over the baseline RLVR methods.

\subsection{Comparison with State-of-the-Art Approaches (RQ3)}

Table~\ref{tab:main_results} presents the comparison with existing multimodal reasoning approaches on five benchmarks using Qwen-2.5-VL-7B as the base model.
TGRL consistently improves both GRPO and DAPO across different training scales, indicating that trajectory-level guidance effectively facilitates multimodal RLVR optimization. 
With 39K training samples, TGRL-DAPO achieves the best average score of 60.68, outperforming state-of-the-art methods. 
Notably, our trajectories are generated by Qwen-2.5-VL-32B, whereas methods like Perception-R1 rely on stronger proprietary models (\textit{e.g.}, Gemini 2.5 pro) for supervision, highlighting the effectiveness of our approach even with comparatively weaker expert models.

\begin{table*}[t]
\caption{
Comparison with state-of-the-art multimodal reasoning methods on five benchmarks using Qwen-2.5-VL-7B as base model. 
TGRL consistently improves both GRPO and DAPO across various training scales and achieves the best performance. 
}
\label{tab:main_results}
\centering
\small
\setlength{\tabcolsep}{1pt}
\begin{tabular}{l c c c c c c c}
\toprule
Method & Data & WeMath & MathVista & MathVerse & MathVision & HalluBench & Avg. \\
\midrule
VLAA-Thinker \cite{vlaa_thinking} & 25K & 67.90 & 68.80 & 49.90 & 26.90 & 68.60 & 56.42 \\
Perception-R1 \cite{perception_r1} & 1.4K & 72.00 & 74.20 & 54.30 & 28.60 & - & - \\
ThinkLite \cite{thinklite} & 11K & 69.20 & 72.70 & 50.20 & 27.60 & 71.00 & 58.14 \\
NoisyRollout \cite{NoisyRollout} & 6.4K & 70.30 & 74.50 & 53.0 & 30.60 & 72.20 & 60.12 \\
R1-Onevision \cite{r1_onevision} & 165K & 62.10 & 63.90 & 46.10 & 22.50 & 65.60 & 52.04 \\
OpenVLThinker \cite{openvlthinker} & 50K & 67.80 & 71.50 & 48.00 & 25.00 & 70.80 & 56.62 \\
\midrule
Qwen-2.5-VL-7B & - & 63.10 & 67.50 & 46.20 & 25.00 & 64.60 & 53.28 \\
\midrule
+ GRPO & 2.1K & 67.40 & 70.50 & 50.80 & 27.30 & 69.80 & 57.16 \\
\rowcolor{blue!15}+ TGRL-GRPO & 2.1K & 70.29 & 71.50 & 51.08 & 27.41 & 72.02 & 58.46\\
+ GRPO & 39K & 68.20 & 70.90 & 50.46 & 28.32 & 70.66 & 57.71 \\
\rowcolor{blue!15} + TGRL-GRPO & 39K & 70.23 & 73.20 & 52.50 & 29.87 & 71.08 & 59.38\\
\midrule
+ DAPO & 2.1K & 69.30 & 70.30 & 50.60 & 26.50 & 67.90 & 56.92 \\
\rowcolor{blue!15} + TGRL-DAPO & 2.1K & 71.05 & 71.40 & \textbf{55.21} & 28.61 & 71.29 & 59.51\\
+ DAPO & 39K & 68.84 & 71.30 & 50.80 & 28.94 & 70.87 & 58.15 \\
\rowcolor{blue!15} + TGRL-DAPO & 39K & \textbf{72.20} & \textbf{75.00} & 53.51 & \textbf{31.26} & \textbf{72.93} & \textbf{60.68} \\
\bottomrule
\end{tabular}
\end{table*}
\section{Related Works}
\label{sec:related_works}
We review related work along three directions: \textbf{(1)} RLVR-based multimodal reasoning, \textbf{(2)} perception-augmented RLVR, and \textbf{(3)} trajectory-guided strategies. We then situate our TGRL within this landscape and clarify its distinctions.

\subsection{RLVR for Multimodal Reasoning}
Recent advancements~\cite{Adavip, lu_mm, dama, lu2024rethinking, lu_ToR,diffgad}, particularly the integration of RLVR, have significantly improved the reasoning capabilities of MLLMs. 
In particular, group-based policy optimization methods such as GRPO~\cite{deepseekmath} and DAPO~\cite{dapo} stabilize reinforcement learning by computing relative advantages within groups of sampled responses.
Building upon these foundations, several works directly apply large-scale RLVR to multimodal models, showing that verifiable rewards can elicit chain-of-thought reasoning and self-correction behaviors without additional supervision~\cite{visual_rft, NoisyRollout,durian}.

Beyond direct RL optimization, some approaches incorporate agentic tool use to actively manipulate visual inputs and interleave image–text interactions during reasoning.
By enabling iterative visual grounding and refinement, works like~\cite{thyme, deepeyes} enhance the model’s ability to utilize visual information.
Others leverage reasoning priors from reasoning LLM backbones, such as the Skywork-R1V series~\cite{skywork_r1v, skywork_r1v2}, or adopt cold-start strategies that warm up models with high-quality SFT data before large-scale RLVR~\cite{vision_r1, cold_start}.
These works primarily focus on improving reasoning capacity under outcome-level sparse reward.

\subsection{Perception-Augmented RLVR}
A key challenge in multimodal reasoning is the perception–reasoning gap, where visual grounding errors propagate into downstream reasoning failures.
To mitigate this issue, perception-augmented RL methods explicitly strengthen the visual grounding stage.

Existing approaches generally fall into two categories.
\textbf{(1)} Some methods introduce stochastic perturbations or transformation strategies during training to improve visual robustness~\cite{NoisyRollout, PAPO};
\textbf{(2)} Others design auxiliary perception-level rewards to encourage faithful grounding before engaging in higher-level reasoning~\cite{perception_r1}.
However, these methods primarily focus on improving perception quality, while relying on sparse outcome-level rewards for the subsequent reasoning process, leaving the perception-to-reasoning transition largely unexplored.

\subsection{Trajectory-Guided RLVR}
Trajectory-guided reinforcement learning introduces expert trajectories into the policy optimization process to enhance the reasoning capability of the current policy.
For text-only LLMs, works such as~\cite{luffy, exgrpo, rlep} incorporate successful reasoning trajectories into the RL loop, stabilizing the reasoning distribution while preserving policy-gradient optimization.

However, extending trajectory guidance to multimodal reasoning remains challenging.
Perception–reasoning coupling amplifies error propagation, while expert trajectory injection inevitably introduces off-policy distribution mismatch.

\subsection{Differences} 
Unlike perception-oriented methods that primarily enhance visual grounding, TGRL explicitly supervises the transition from perception to reasoning through trajectory-level alignment. 
Moreover, compared to prior trajectory-guided RL approaches developed for text-only settings, TGRL adapts expert trajectory injection to multimodal RLVR by carefully handling perception–reasoning coupling and off-policy distribution mismatch. 
Overall, TGRL extends group-based multimodal RL optimization from outcome-level supervision to trajectory-level alignment, integrating expert guidance while preserving on-policy exploration.
\vspace{-1ex}
\section{Conclusion}
\label{sec:conclusion}
In this work, we revisit the optimization flow of multimodal RLVR and identify the lack of supervision over the perception–reasoning transition as a critical bottleneck. 
To address this, we propose \textbf{T}rajectory-\textbf{G}uided \textbf{R}einforcement \textbf{L}earning (TGRL), which integrates expert trajectories into group-based RL optimization through token reweighting and trajectory filtering. 
By extending supervision from outcome-level rewards to trajectory-level alignment, TGRL enables more precise credit assignment while preserving on-policy exploration.

\noindent \textbf{Limitations and Future Work.}
Future work includes reducing reliance on strong expert policies through self-generated trajectories and improving the stability of off-policy trajectory alignment. 
Extending trajectory-level supervision to broader multimodal settings would further validate its generality and advance more reliable perception–reasoning integration in multimodal foundation models, particularly for embodied intelligence and real-world reasoning tasks.

\clearpage  


%
%
\bibliographystyle{splncs04}
\bibliography{main}
\end{document}